# On-Device Continual Learning with Dual-Stage Buffer and Dynamic Loss for Point-of-Care Pneumonia Diagnosis

Danu Kim
Korea International School, Jeju Campus, Jeju-do 63644, Republic of Korea (e-mail: dukim27@kis.ac)

**(Abstract) Deep learning models detect pneumonia from chest X-rays with high accuracy, but the performance declines under domain shifts caused by differences in devices, patients, or institutions. We present *PneumoNet*, a domain-incremental learning method for point-of-care pneumonia diagnosis in resource-limited settings. PneumoNet combines a lightweight CNN for on-device prediction, a dual-stage balanced buffer for class-balanced replay, and a dynamic class-weighted loss to correct training-batch imbalances. Evaluated on a domain-shifted PneumoniaMNIST dataset simulating five realistic domain change scenarios, PneumoNet achieves 86.6% accuracy with 1.4% forgetting while being smaller and faster than existing baselines. These results highlight PneumoNet's potential to enable adaptive, privacy-preserving diagnostic AI directly on point-of-care medical devices in real-world and pandemic-ready healthcare.**

## 1. INTRODUCTION

Since the discovery of X-rays in 1895 by Wilhelm Conrad Roentgen, radiographic imaging changed clinical diagnosis and patient management [1, 2]. In chest radiography, early detection of pneumonia is critical for effective treatment, particularly in vulnerable patients such as children. Until now, various deep learning methods have achieved expert-level performance in pneumonia detection from large-scale public chest X-ray datasets such as CheXpert [3], MIMIC-CXR [4], and PneumoniaMNIST [5]. However, these approaches are typically developed under the assumption that the training and test data share the same distribution, a condition rarely met in real-world medical environments.

In practice, domain shifts/variations in X-ray imaging are inevitable. They arise from device upgrades, protocol changes, patient demographics, or institutional variations, leading to degraded model performance. Retraining the models centrally and completely using all the aggregated multi-domain data is one solution, but this is impractical due to patient privacy regulations, limited resources, especially in on-device clinical settings that require local updates.

Mobile X-ray machines, designed for portability, are essential in intensive care units, emergency departments, or remote and resource-limited settings. However, their wide use in diverse environments increases domain shifts of X-ray data, caused by variations in calibration, imaging angles, lighting conditions, and patient positioning. With respiratory diseases expected to trigger the next pandemics, rapid pneumonia identification via mobile X-ray systems will be critical for large-scale screening and triage. In such scenarios, mobile X-ray systems will function as essential point-of-care diagnostic devices, enabling timely and accessible pneumonia detection in resource-constrained environments. This research focuses on pneumonia detection as a respiratory disease challenge, aiming to ensure that diagnostic AI models can adapt quickly and reliably to evolving conditions without external retraining, a prerequisite for resilient and globally deployable healthcare systems.

Recently, continual learning (CL) methods have emerged as a promising way for adapting models without catastrophic forgetting of prior knowledge. CL is typically categorized into three types: task-incremental (new tasks with task identity provided), class-incremental (new classes are introduced over time), and domain-incremental (the task remains the same, but the input distribution changes over time).

Among these, domain-incremental learning (DIL) is the most relevant for medical imaging, as the diagnostic task remains unchanged while data distributions shift. Although prior CL studies in medical imaging have addressed task- and class-incremental settings, applications of DIL to chest X-ray analysis, particularly pneumonia detection, are still limited. Replay-based continual learning that store and selectively replay a small set of past samples have consistently achieved the strongest performance across CL benchmarks. Motivated by this, we adopt replay as the main part of our method.

A further challenge in medical image detection tasks is class imbalance. Many clinical datasets contain much more normal cases than positive disease cases, or the opposite, depending on the experimental setup. This imbalance biases the model toward majority classes, degrading sensitivity for rare but clinically critical cases. In continual learning, the problem becomes more severe when new domains introduce further imbalance and replay buffers are small. Thus, effective DIL method must explicitly address class imbalance during both replay sampling and model optimization. Our proposed method uses dual-stage balanced buffer to maintain class-balanced replay and a dynamic class-weighted loss function to adaptively address within each batch, preserving diagnostic performance for minority classes.

To the best of our knowledge, this is the first study to apply DIL to X-ray pneumonia diagnosis. For systematic evaluation, we created a custom domain-shifted PneumoniaMNIST using controlled transformations that simulate realistic clinical variations; LowDose (reduced radiation), Portable (brightness and blur changes for portable ICU imaging), Anatomical (translation and scaling for patient posture or body type differences), and Institutional (contrast, brightness, and sharpness changes for hospital-to-hospital variations).

We then propose PneumoNet, a lightweight on-device DIL method designed for resource limits while maintaining robust performance. While replay-based methods are widely used in continual learning, their high computational and memory demands make on-device deployment difficult. PneumoNet handles these challenges through efficient convolutional neural network (CNN) architecture combined with the dual-stage balanced buffer and dynamic class-weighted loss.

The contributions of this work are threefold:

- First application of domain-incremental learning to X-ray pneumonia detection, through the creation of a domain-shifted PneumoniaMNIST dataset for systematic evaluation.
- Design of an efficient continual learning method, integrating lightweight PneumoNet for fast inference, dual-stage balanced buffer for class-balanced replay, and class-weighted loss for intra-batch imbalance correction.
- Extensive experiments showing high accuracy, low forgetting, and strong efficiency across five sequential domains under the resource constraints representing mobile and portable medical devices.

## 2. Related Work

### 2.1 Chest X-ray Datasets

Over the past decade, numerous chest X-ray (CXR) datasets have been released, each designed to address specific research needs in medical imaging. Early collections such as Demner-Fushman et al. provided 8,121 images with carefully curated reports, demonstrating the feasibility of structured radiology corpora for machine learning [6].

Scaling up, ChestX-ray8 introduced over 100,000 images labeled for eight thoracic diseases, setting a benchmark for large-scale disease classification [7]. Datasets like CheXpert [3] and MIMIC-CXR [4] extended this work, providing over 220,000 images and 370,000 images respectively from a vast number of studies and associated free-text radiology reports, making them central to modern deep learning in radiology.

More specialized datasets have been designed to target pneumonia and other lung conditions. Shih et al. provided bounding box annotations for pneumonia-like opacities [8], enabling localization tasks, while Filice et al. provided crowd-sourced annotations for pneumothorax, demonstrating hybrid AI–human labeling [9]. During the COVID-19 pandemic, Cohen et al. compiled an international CXR dataset for rapid AI deep learning development [10].

To increase demographic and clinical diversity, PadChest introduced a multilingual dataset with over 160,000 images [11], while VinDr-CXR [12] and PediCXR [13] focused on high-quality annotations in Vietnam, including pediatric populations that are often underrepresented in prior datasets.

Finally, lightweight benchmark collections such as MedMNIST [5] divided large biomedical datasets into small, standardized subsets for rapid benchmarking. Among them, PneumoniaMNIST, based on 5,856 pediatric CXRs, provides an accessible platform for binary classification of pneumonia versus normal cases and forms the foundation of our domain-shifted experiments.

### 2.2 X-ray Deep Learning

These datasets enabled rapid advances in deep learning for CXR analysis. Rajpurkar et al. demonstrated the potential of deep convolutional networks by developing CheXNet [14], achieving radiologist-level pneumonia detection on ChestX-ray14 dataset. Variants such as fine-tuned VGG16 [15] outperformed state-of-the-art deep learning algorithms such as ResNet-50 in pneumonia classification task, and Xception-based models [16] extended this approach to broader thoracic findings, achieving performance comparable to board-certified radiologists.

The COVID-19 pandemic expanded the use of transfer learning approaches. For example, Apostolopoulos et al. [17], Ucar et al. [18], and Abbas et al. [19] applied CNNs and model compression methods to COVID diagnosis, achieving accuracy above 90%. Minaee et al. further demonstrated high sensitivity and specificity using pre-trained CNNs [20]. Albahli et al. extended the approach by applying pre-trained CNNs to classification of COVID-19 and 14 other chest diseases [21].

In addition to individual-study models, libraries such as TorchXRayVision [22] provided easy access to datasets and pre-trained models, making benchmarking and generalization studies convenient. Lightweight designs such as Yen et al. [23] showed that efficient CNNs could match or outperform larger architecture, supporting mobile deployment. Overall, these studies demonstrate the strong potential of deep learning for pneumonia detection, but also highlight the challenges in generalization, especially when models trained on one dataset are tested on unseen data.

### 2.3 Challenges and Bias in Chest X-ray AI

A critical limitation of current CXR AI is dataset bias and poor cross-domain generalization. Cohen et al. [24] and Liu et al. [25] pointed out that inconsistent labeling, inter-observer variability, and spurious correlations often lead to accuracy drops in model performance.

Empirical studies have confirmed these disparities. Glocker et al. found sex/race-specific accuracy drops of up to 11% in foundation models [26], while Kobayashi et al. demonstrated systematic underdiagnosis in underrepresented groups due to biases in data generation process [27], including variations in disease prevalence, clinical workflow, and image acquisition. These findings emphasize the importance of approaches like ours that explicitly address domain shifts and ensure fairness across diverse populations.

The integration of continual learning into clinical practice raises both opportunities and risks. Lee et al. cautioned that dynamically updating models can directly influence patient outcomes yet lack reliable validation frameworks [28]. The U.S. FDA recognized this challenge [29] and proposed regulatory mechanisms such as "predetermined change-control plans" to allow adaptive models while reducing the risks of bias, catastrophic forgetting, or unsafe updates. This changing regulatory environment needs for continual learning methods that are transparent, efficient, and safe.

### 2.4 Continual Learning for Medical Imaging

Continual learning research has developed many strategies for forgetting problems. Elastic Weight Consolidation (EWC) penalizes updates to critical parameters [30], while iCaRL combines exemplar storage with nearest-mean classification [31]. Memory-based methods such as GEM [32], A-GEM [33], and MIR [34] rely on episodic replay buffers to control gradient updates.

Experience Replay (ER) remains a strong baseline [35], with studies showing that even small episodic memories outperform more complex methods in many cases. Extensions such as GSS [36], CBRS [37], and reservoir sampling [38] further improve memory efficiency and class balance. Frameworks like Avalanche [39] and taxonomies by Van de Ven et al. [40] has standardized evaluation across task-, class-, and domain-incremental learning. Overall, replay-based CL methods continue to show strong performance, but their high computational and memory costs make them less practical for on-device deployment, a key motivation for our work.

Applying CL to medical imaging is still emerging but rapidly expanding. Baweja et al. applied EWC to Brain MRI segmentation [41], while Derakhshani et al. benchmarked CL methods on MedMNIST, showing iCaRL's competitiveness in task-incremental case [42]. Verma et al. evaluated privacy-preserving CL for optical coherence tomography images, highlighting Brain-Inspired Replay as a promising exemplar-free approach [43].

Other applications include ultrasound image classification with ResNet-64 [44], cardiac MRI segmentation with ResNet-50 [45], and multi-organ segmentation with Lifelong nnU-Net [46]. Surveys such as Li et al. have reviewed CL applications to physiological signals, from ECG to EEG [47]. These works demonstrate the feasibility of CL across medical data types and show that most studies focus on task or class-incremental settings. Applications of domain-incremental learning, where the task remains fixed, but the input distribution shifts, remain scarce. Our study directly addresses this gap in the context of pneumonia detection from chest X-rays.

### 2.5 Summary and Research Gap

The literature review presents three main trends: (1) CXR datasets have become larger and more diverse, enabling deep learning models with radiologist-level accuracy; (2) despite these progresses, cross-domain generalization remains limited by dataset bias, demographic and institutional variability; (3) continual learning shows promise in medical imaging, yet most previous studies leave domain-incremental CL underexplored.

Together, these findings show a critical gap: deep learning models perform well under fixed distributions but struggle with domain shifts common in clinical practice. Existing continual learning methods often depend on task labels or heavy replay, limiting use on resource-constrained devices. To address this, we propose a lightweight domain-incremental deep learning method developed for on-device pneumonia detection.

## 3. PNEUMONET: PROPOSED METHOD

### 3.1 Approach Overview

To our knowledge, this work is the first to explore domain-incremental continual learning for X-ray pneumonia detection. While prior continual learning studies in medical imaging have studied class-incremental or task-incremental scenarios, they have not systematically tackled the challenge of adapting to domain shifts in X-ray images such as those arising from changes in imaging devices, scan protocols, patient groups, or hospital-specific calibration settings, while still preserving diagnostic accuracy on earlier domains. This gap is critical, as real-world hospital environments often undergo such domain changes, making it essential to develop learning strategies that adapt without catastrophic forgetting.

Our approach is designed for on-device deployment in low-resource clinical settings, where full central retraining is often impractical due to privacy, speed, and network issues. We propose a lightweight continual learning method that integrates three core ideas:

1. **Efficient CNN Architecture:** a lightweight CNN model optimized for real-time use on mobile/portable devices, reducing memory use, FLOPs, and inference time while keeping high accuracy.
2. **Dual-Stage Balanced Buffer:** a memory management strategy combining per-class reservoir sampling with balanced replay selection, ensuring balance within the buffer's content and maintains balance in the replayed batch from the buffer
3. **Dynamic Class-Weighted Loss:** an adaptive class loss reweighting method that adjusts for intra-batch class imbalance, preventing the dominance of majority classes and improving minority-class performance.

Fig. 1 shows the continual learning process of the proposed method, where the neural model is continually updated as new domains from different hospitals, patients are introduced. At each incremental step, incoming X-ray samples from a new domain are partially stored in the replay buffer according to per-class reservoir rules. Balanced replay samples from past domains are then combined with current-new domain data. A dynamically weighted loss is computed to deal with class imbalance, and model parameters are updated incrementally. This approach allows the model to learn domain-specific variations in pneumonia appearance while keeping prior diagnostic knowledge, all within mobile X-ray computational and memory limits.

The remainder of this section is organized as follows: Section 3.1 describes the domain-shifted dataset construction, including the transformation and simulations for each domain; Section 3.2 explains the PneumoNet architecture for efficient on-device inference; Section 3.3 introduces the dual-stage balanced buffer for balanced replay; Section 3.4 presents the dynamic class-weighted loss for adaptive balancing; and Section 3.5 describes the overall training algorithm of the proposed method.

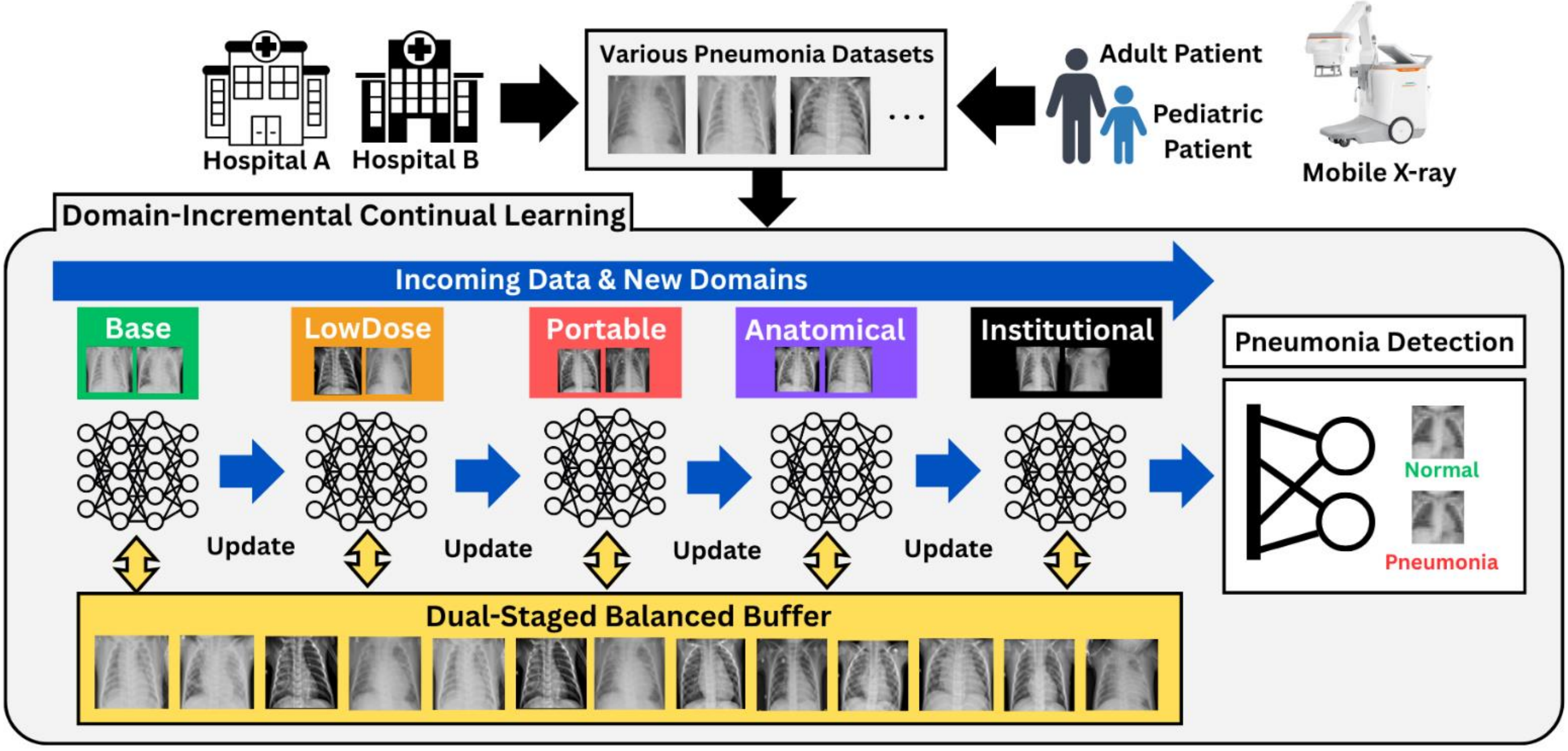

**Fig. 1 Overview of the proposed domain-incremental learning method for pneumonia detection from chest X-rays.** The model is continually updated as new domains (representing different imaging sources or clinical environments) are introduced. A lightweight CNN model with dual-stage balanced buffer stores representative samples from previous domains, enabling replay-based training that mitigates catastrophic forgetting and maintains consistent diagnostic performance across evolving imaging conditions.

### 3.2 Domain-Shifted Dataset Construction

A major challenge in domain-incremental learning (DIL) research is the absence of public datasets that capture progressive, clinically relevant domain shifts. To overcome this limitation, we construct a custom domain-shifted version of PneumoniaMNIST by applying controlled transformations that mimic real-world acquisition and demographic variations. These include low-dose noise injection, contrast and brightness adjustments, blurring to simulate portable imaging artifacts, and anatomical modifications reflecting differences in posture or body habitus. The resulting dataset provides a benchmark for evaluating DIL methods under realistic clinical shifts when public resources are unavailable.

Because this study represents, to our knowledge, the first application of DIL to pneumonia detection, a domain-shifted dataset is essential. Accordingly, we extend the base PneumoniaMNIST dataset into four additional domains, each designed to simulate a specific real-world scenario, as described below.

1. **Base:** Original PneumoniaMNIST images.
2. **LowDose:** Noise injection and intensity reduction to simulate low-dose radiation imaging.
3. **Portable:** Slight brightness changes and blurring to simulate the ICU imaging with uneven lighting and patient positioning.
4. **Anatomical:** Translations and scaling to represent variations in posture and body habitus such as pediatr ic or obese patients.
5. **Institutional:** Changes in contrast, brightness levels, and sharpness to mimic inter-hospital variability caus ed by different equipment and imaging protocols.

### 3.3 PneumoNet Architecture for On-Device Prediction

On-device applications of diagnostic AI require models that are highly efficient in size, parameters, memory use, and computational cost. Large deep neural networks are often impractical for mobile deployment, as they require excessive floating-point operations, training and inference time. To deal with these constraints, we design PneumoNet, a lightweight convolutional neural network optimized for pneumonia prediction under resource limits.

Fig. 2 illustrates the PneumoNet architecture for learning diagnostic features from portable CXR images. The feature extraction stage comprises two convolutional blocks that progressively capture spatial patterns relevant to pneumonia detection. The first block uses 3×3 kernels with 16 feature maps to detect basic structures such as edges and lung boundaries, followed by ReLU activation and 2×2 max pooling to preserve key features while reducing the image from 28×28 to 13×13. The second block extracts more complex textures and opacity patterns associated with infection, producing 32 feature maps of size 5×5 after pooling. This hierarchical extraction enables efficient representation learning with low computational cost, suitable for mobile diagnostic systems.

In the classification stage, the resulting feature maps are flattened into an 800-dimensional vector and passed through a fully connected layer with 32 ReLU units. The final layer outputs two logits corresponding to normal and pneumonia, which are converted to probabilities using softmax operation for interpretable diagnosis. Overall, this lightweight CNN provides a fast and accurate framework for pneumonia detection in low-resource clinical and mobile healthcare environments.

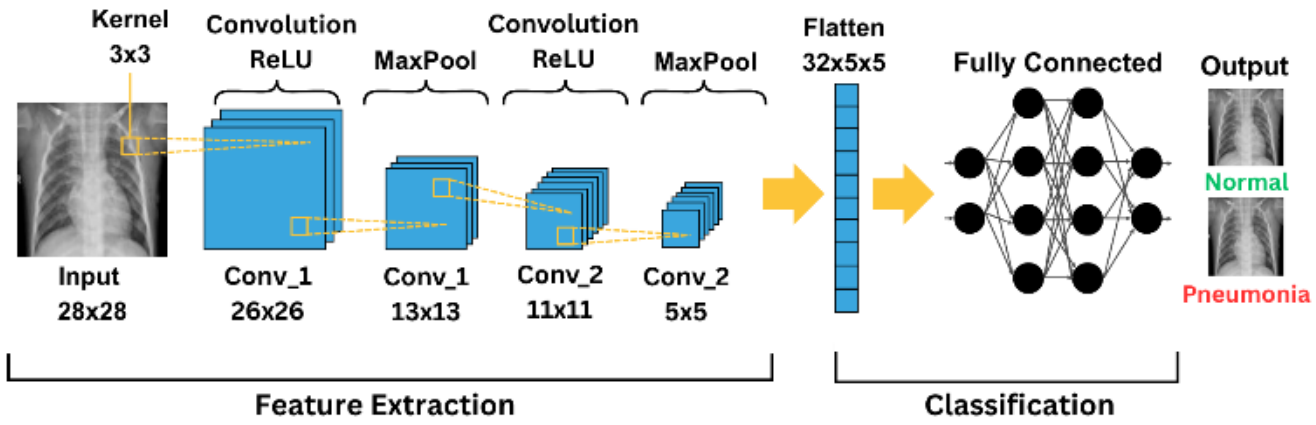

**Fig. 2. PneumoNet Architecture.** It consists of a feature extraction using two convolutional-pooling blocks and a classification stage.

This design offers two main advantages. First, PneumoNet has a small model size and low memory usage, with far fewer parameters and storage requirements than larger networks. Second, its shallow depth enables fast inference, enabling real-time predictions on edge devices with limited computing power. Together, these features allow PneumoNet to balance diagnostic accuracy with the efficiency required for continual learning in mobile X-ray systems.

### 3.4 Dual-Stage Balanced Buffer for Class Balancing

Medical datasets often have imbalanced classes, such as many more normal chest X-rays than pneumonia cases. In continual learning, this problem becomes even worse because the replay buffer is small, and each new domain may have a different class ratio. Conventional reservoir sampling method usually keeps a fixed memory by uniformly sampling from the input data, but this does not handle imbalance. As a result, training batches are dominated by majority classes, so the model updates mainly focus on majorities and overlook the rare but important cases like pneumonia.

To address this issue, we propose a dual-stage balanced buffer that maintains data balance from two perspectives (See Algorithm 1). Firstly, the *Add_Batch* stage balances the buffer content when storing new samples using per-class reservoir storage, ensuring each class keeps a fair share of memory. The *Get_Sample* stage then rebalances during replay by sampling equally from each class, preventing minority classes from being overlooked during training.

**Add_Batch:** it stores new image samples in the class-specific buffers for pneumonia data and normal data.

1. If the target buffer $buffer[cls = y_i]$ for class $y_i$ has fewer than $K$(capacity per class) samples, the incoming example $(x_i, y_i)$ is appended.
2. Once full, a reservoir-sampling policy is applied: each new $(x_i, y_i)$ randomly replaces an existing one in the same buffer, ensuring fair representation.

**Get_Sample:** it retrieves samples from the buffer for training. It first identifies all classes with at least one stored sample in the buffer.

1. If none exists, it returns $(None, None)$.
2. Otherwise, it calculates the per-class quota:
$$k = M/C$$
where $M$ is the replay size and $C$ is the number of classes.
3. Up to $k$ samples are randomly selected from each available class buffer and added to the replay set.

---

**Algorithm 1** Dual-Stage Balanced Buffer

**Input:** Capacity per class $K$; number of classes $C$; replay size $M$
**Output:** Updated buffer and replay set

```
1:  procedure ADD_BATCH(input: (x_1, y_1), ..., (x_n, y_n))
2:      for each sample (x_i, y_i) do
3:          Identify class cls ← y_i
4:          if buffer[cls].len < K then
5:              Append (x_i, y_i) to buffer[cls]
6:          else
7:              Randomly replace in buffer[cls] with (x_i, y_i)
8:          end if
9:      end for
10: end procedure
11: procedure GET_SAMPLE(input: replay size M)
12:     Identify classes for which buffer[cls] is non-empty
13:     if no such classes exist then
14:         return (None, None)
15:     end if
16:     Compute the number of replay samples per class:
                    k ← M/C
17:     for each cls in non-empty classes do
18:         Add k random samples from buffer[cls] to replay_set
19:     end for
20:     if replay_set.len < M then
21:         Fill the remainder with random samples from all buffers
22:     end if
23:     return replay_set
24: end procedure
```

---

4. If the total remains below $M$ (e.g., due to under-filled buffers), the remaining slots are filled with random samples from all stored samples.

By enforcing balance at both the storage and replay stages, the dual-stage buffer reduces the dominance of majority-class, stabilizes continual model learning over time, and preserves diagnostic accuracy across sequential domains variations.

### 3.5 Dynamic Class-Weighted Loss

Even with a balanced replay buffer, imbalance may still appear in the combined training batch, since incoming data from the current domain can be skewed. If not corrected, the model may overfit to the majority class within each training batch.

To address this, we introduce a dynamic class-weighted loss that adaptively reweighs the contribution of each class during training. Let $n$ denote the number of samples in a batch and $C$ the number of classes. For the $i$-th class, let $n_i$ be the number of samples of that class in the batch. We define the class weight as:

$$W_i = \frac{n \cdot C}{n_i}$$

This assigns higher weights to underrepresented classes and lower weights to majority classes.

For example, consider a batch with $n = 10$ samples and $C = 2$ classes (pneumonia and normal). Suppose the batch contains $n_{\text{pneu}} = 2$ and $n_{\text{norm}} = 8$. Then, the corresponding class weights are:

$$W_{\text{pneu}} = \frac{10 \cdot 2}{2} = 10$$

and

$$W_{\text{norm}} = \frac{10 \cdot 2}{8} = 2.5$$

In this case, the minority class (pneumonia) receives a larger weight than the majority class (normal)

Then, the class-weighted cross-entropy Loss is computed as:

$$L_{weighted} = -\frac{1}{n}\sum_{j=1}^{n} W_{y_j} \cdot \log p_j$$

where $y_j$ is the true label of input sample $j$ and $p_j$ is the predicted probability of that true class.

This dynamic reweighting ensures that rarer classes, such as pneumonia in many datasets, have proportionally greater influence during optimization with higher weight values. By adjusting automatically to the composition of each training batch, the weighted loss, together with the dual-stage buffer, further reduces class imbalance by adapting to class frequency variations, improving minority-class prediction over domain shifts.

### 3.6 Training Algorithm

We now describe the proposed training algorithm (See Algorithm 2), which generalizes the training process of PneumoNet. The training strategy operates in a sequential continual learning setting, where data batches arrive one after another from an input $\{(x_1, y_1), \ldots, (x_n, y_n)\}$. The learner maintains a neural model $f(x)$ and a replay buffer that stores a subset of previously seen samples. Training proceeds over multiple epochs, with each iteration processing both new and replayed data.

In each epoch, the model first reads a batch $(x_{new}, y_{new})$ from input. It then retrieves replay samples $(x_{replay}, y_{replay})$ from the buffer, with a size proportional to the replay ratio (default = 1.0, meaning the replay set and new input have equal sizes). The dual-stage balanced buffer's *Get-Sample* method is used to maintain class balance. Finally, the two sets are combined into a single batch $(x_{combined}, y_{combined})$.

The combined batch is passed through PneumoNet to produce predictions $\hat{y} = f(x_{combined})$. Because PneumoNet is a lightweight CNN, inference remains efficient. The class-weighted loss is then computed using the class-specific weights to mitigate the class imbalance during optimization. The model parameters are updated to reduce this loss through backpropagation.

Finally, the current batch $(x_{new}, y_{new})$ is stored to the buffer using the *Add_Batch* method. This process repeats until the specified number of epochs is completed, enabling continual learning across sequential domains while limiting forgetting.

**Algorithm 2** Proposed Training Algorithm

**Input:** Input data $\{(x_i, y_i)\}_{i=1}^{n}$, model $f(\cdot)$, loss function $\mathcal{L}(y, \hat{y})$
**Output:** Trained model parameters

1: **repeat**
2: Read a batch $(x_{\text{new}}, y_{\text{new}})$ from the input.
3: Retrieve $(x_{\text{replay}}, y_{\text{replay}})$ from buffer using GET_SAMPLE.
4: Combine current and replayed data:

$$(x_{\text{combined}}, y_{\text{combined}}) = (x_{\text{new}} \cup x_{\text{replay}},\ y_{\text{new}} \cup y_{\text{replay}})$$

5: Predict outputs: $\hat{y} = f(x_{\text{combined}})$.
6: Compute class-weighted loss:

$$L_{weighted} = -\frac{1}{n}\sum_{j=1}^{n} W_{y_j} \cdot \log p_j$$

7: Update model parameters to minimize $\mathcal{L}$.
8: Store new samples $(x_{\text{new}}, y_{\text{new}})$ in buffer using ADD_BATCH.
9: **until** number_epochs is completed

## 4. EXPERIMENTS

### 4.1 Experimental Setup

**Datasets.** We evaluate PneumoNet on DIL using custom domain-shifted PneumoniaMNIST. Original dataset contains 5,856 pediatric chest X-ray images (28x28 pixel) for binary classification of pneumonia versus normal. Our domain-shifted version consists of five domains described in section 3.2: Base, LowDose, Portable, Anatomical, and Institutional. Each domain simulates real-world variations including noise, blurring, anatomical variations, and variations in contrast or brightness. Evaluation follows the sequential order: Base → LowDose → Portable → Anatomical → Institutional.

**Models and Hyperparameters.** We compare the proposed method with two popular CL methods: Experience Replay (ER) and Class-Balancing Reservoir Sampling (CBRS). ER is the best baseline method and benefits from reservoir sampling, while CBRS explicitly balances buffer contents for each class. For both methods, the replay buffer size is limited to 500. In addition, we compare Joint-Training (upper bound) and Fine-Tuning (lower bound), trained on the complete dataset. All models are trained with Adam optimizer (learning rate = 0.001), batch size = 32, and 50 epochs per domain.

**Metrics.** Performance is evaluated by the following metrics, and results averaged over three runs were reported.

1. *Average Accuracy*: Mean test accuracy across all domains after CL.
2. *Average Forgetting*: Average gap between the highest and final accuracy of each domain.
3. *Model Size*: File size including weights and architecture.
4. *Memory Usage*: RAM or GPU memory during inference
5. *Parameters*: Number of trainable weights and biases.
6. *FLOPs*: Floating-point operations per forward pass.
7. *Training Time*: Time required to train the model.
8. *Inference Time*: Time per batch to generate predictions.

**Table 1. Performance comparison of PeumoNet and baselines in DIL continual learning.** (The best results are highlighted in bold.)

| Method | Accuracy (%) | Forgetting (%) |
|---|---|---|
| Joint-Training (upper bound) | 87.17 (± 1.34) | 0.00 (± 0.00) |
| Fine-Tuning (lower bound) | 80.92 (± 0.61) | 5.02 (± 0.74) |
| ER | 84.10 (± 0.94) | 3.45 (± 1.19) |
| CBRS | 84.96 (± 1.14) | 2.74 (± 0.38) |
| **PneumoNet** | **86.58 (± 0.36)** | **1.43 (± 0.24)** |

**Table 2. Performance comparison of PneumoNet on different domains during continual learning.**

| Domain | After D1 | After D2 | After D3 | After D4 | After D5 |
|---|---|---|---|---|---|
| Base | 86.33 | 86.38 | 87.50 | 90.06 | 88.73 |
| LowDose | - | 85.79 | 83.55 | 83.92 | 83.55 |
| Portable | - | - | 87.39 | 87.77 | 86.91 |
| Anatomical | - | - | - | 87.98 | 85.47 |
| Institutional | - | - | - | - | 88.25 |

### 4.2 Performance Comparison for Continual Learning

Table 1 compares the accuracy between PneumoNet and baseline models. Joint-Training achieves the highest accuracy (87.17%) with zero forgetting, but it assumes full access to all domain at once, which is unrealistic in continual learning. Fine-Tuning, by contrast, suffers severe forgetting (5.02%) and the lowest accuracy (80.92%). Among continual learning methods, PneumoNet outperforms both ER and CBRS. Compared to ER, PneumoNet improves accuracy by 2.48% and reduces forgetting by 2.02%. In comparison to CBRS, PneumoNet accuracy improves by 1.62% and forgetting by 1.31%. Moreover, PneumoNet shows the smallest standard deviations in both accuracy and forgetting. Despite its lightweight architecture and a small buffer size, PneumoNet closely approaches the Joint-Training upper bound while still maintaining continual-learning constraints.

Table 2 presents PneumoNet's test accuracy across five domains as they are introduced sequentially during continual learning (D1 → D5), averaged over three runs. Base domain shows a stable improvement from 86.33% to 90.06% across the sequence, followed by a slight decrease to 88.73% after Domain 5. For other domains, accuracy declines modestly over time. LowDose accuracy drops from 85.79% down to 83.55%, Portable from 87.39% to 86.91%, and Anatomical from 87.98% to 85.47%.

Despite minor drops, all domains maintain similar accuracy throughout training, showing no signs of severe forgetting. These small declines are likely due to the PneumoNet's compact architecture and limited buffer size, designed for mobile use. As new domains are added, trade-offs between learning new data and retaining old knowledge are expected in continual learning. However, the accuracy decreases by only about 2% in LowDose and Anatomical, and less than 1% in Portable, indicating that forgetting is effectively minimized. Overall, PneumoNet shows stable performance across all domains.

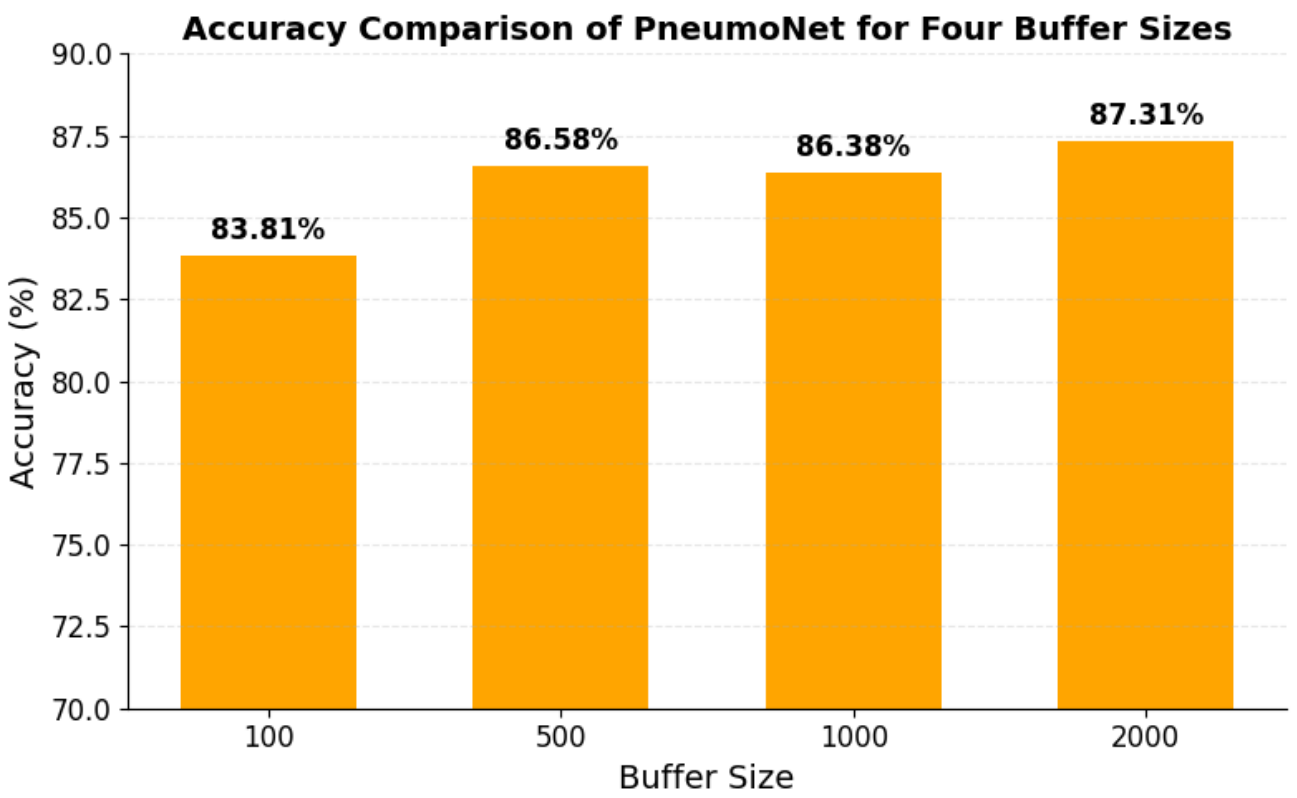

**Fig. 3. Accuracy of PneumoNet for four different buffer sizes.**

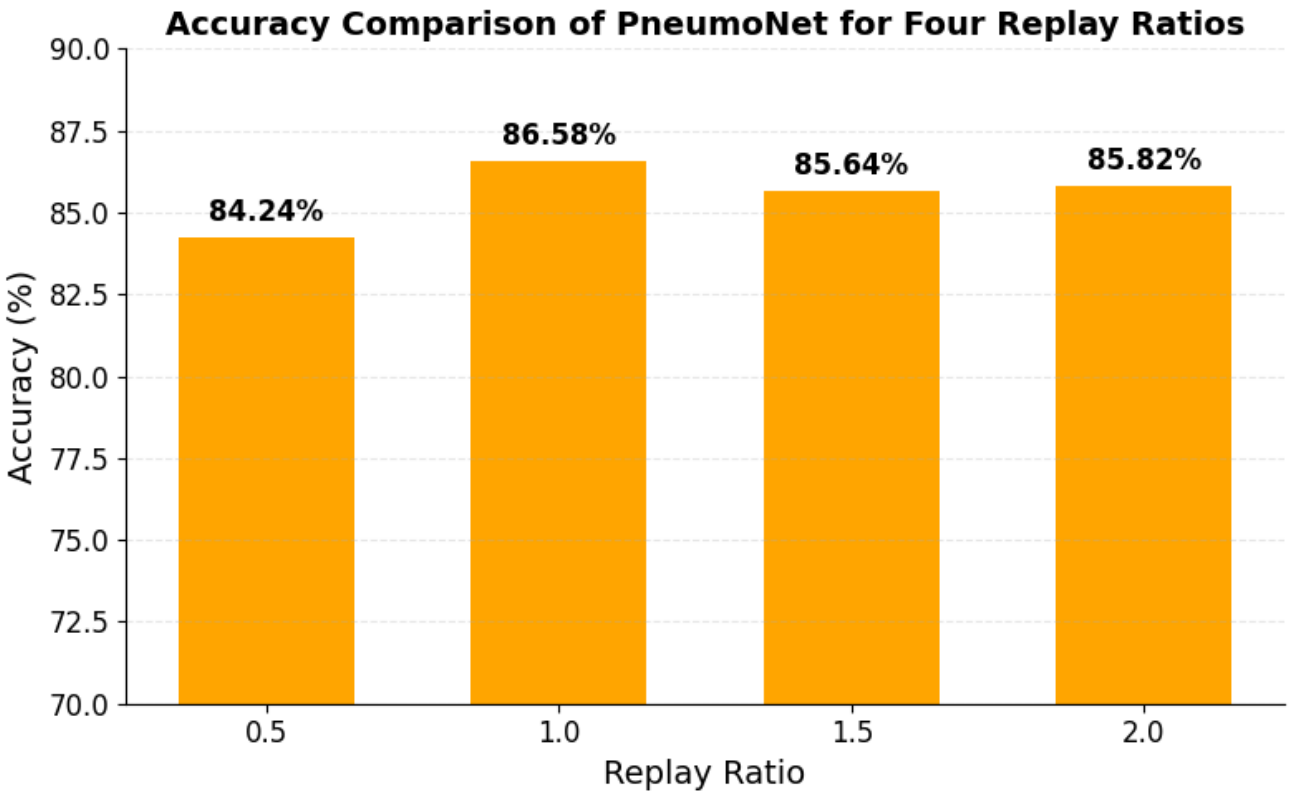

**Fig. 4. Accuracy of PneumoNet for four different replay ratios.**

### 4.3 In-Depth Analysis

To better understand how PneumoNet performs in different situations, we analyzed how factors like different buffer size, replay ratio, optimization strategy, and model type affect its accuracy and forgetting.

Fig. 3 presents the accuracy comparison of the proposed PneumoNet for different buffer sizes (100, 500, 1000, 2000). We change the buffer size to study the trade-off between memory use and accuracy. Larger buffers help the model remember the earlier knowledge better but use more memory, while smaller buffers save memory but may forget past data. This experiment tests whether PneumoNet can still perform well with a small buffer suitable for mobile devices. With a small buffer of 100 replay samples, accuracy reaches 83.81%, showing limited ability to maintain past knowledge. Expanding the buffer to 500 sharply increases accuracy to 86.58%. Increasing further to 1,000 yields a slight drop (86.38%), while 2000 achieves highest accuracy at 87.31%.

These results show that most of the improvement happens when the buffer increases from 100 to 500 samples, while larger buffers provide only small additional gains. Notably, PneumoNet, with a buffer of 500, only one-quarter the size of 2000, already reaches competitive performance. This balance between efficiency and performance makes PneumoNet well suited for mobile X-ray devices that have limited memory.

**Table 3. Performance comparison of PneumoNet based on two optimizers (Adam & SGD).** (The best results are highlighted in bold)

| Optimizer | Accuracy (%) | Forgetting (%) |
|---|---|---|
| Adam | **86.58 (± 0.36)** | 1.43 (± 0.24) |
| SGD | 84.99 (± 0.39) | **0.83 (± 0.57)** |

**Table 4. Performance comparison of PneumoNet based on two training models (CNN vs. MobileNet V2).** (The best results are highlighted in bold.)

| Method | Accuracy (%) | Forgetting (%) |
|---|---|---|
| **PneumoNet** | **86.58 (± 0.36)** | 1.43 (± 0.24) |
| MobileNet-V2 (No-weights) | 83.02 (± 0.49) | 2.88 (± 0.52) |
| MobileNet-V2 (Pretrained) | 86.37 (± 0.42) | **1.09 (± 0.69)** |

Fig. 4 evaluates the average accuracy of PneumoNet for four replay ratios (0.5, 1.0, 1.5, and 2.0). We vary the replay ratio to examine how PneumoNet balances remembering old domains and learning new ones. Too little replay causes the model to forget earlier data, while too much replay slows its ability to learn new information. This experiment checks whether a balanced ratio gives the best results. Replay ratio of 1.0 achieves the best accuracy of 86.58%, while lower replay of 0.5 reduces accuracy to 84.24%. Increasing replay beyond 1.0 leads to lower performance, with accuracies of 85.64% at 1.5 and 85.82% at 2.0. This indicates the importance of balanced replay, equal amounts from current and past data, is optimal. Too little replay limits the model's access to prior knowledge, while excessive replay slows adaptation to new domains. Thus, a replay ratio of 1.0 provides the best trade-off for lightweight continual learning.

Next, we compare Adam and SGD to see how the choice of optimizer affects accuracy and forgetting, identifying which works better for PneumoNet in continual learning. Table 3 shows that the choice of Adam optimizer achieves higher accuracy (86.58% vs. 84.99%) but slightly more forgetting (1.43% vs. 0.83%). This trade-off happens because Adam learns faster and reaches better peak performance, while SGD updates more slowly but helps retain knowledge better. For practical use, Adam is preferred when accuracy is important, while SGD is useful in scenarios where minimizing forgetting is more critical.

Table 4 compares the performance of PneumoNet with two versions of MobileNet-V2 (No-weights vs. Pretrained) to evaluate the changes in accuracy and efficiency. MobileNet-V2 is a well-known lightweight model, while PneumoNet is designed for strict memory limits. This experiment shows if PneumoNet can reach similar accuracy while being much smaller and faster. PneumoNet outperforms MobileNet-V2 (No-weights) trained from scratch (86.58% vs. 83.02%) and performs comparably to the pretrained version (86.37%). Forgetting is lower in pretrained MobileNet-V2 (1.09% vs. 1.43%), but the difference is small. Considering the much smaller size and computational cost of PneumoNet, it offers state-of-the-art performance while staying highly efficient for continual learning on mobile devices.

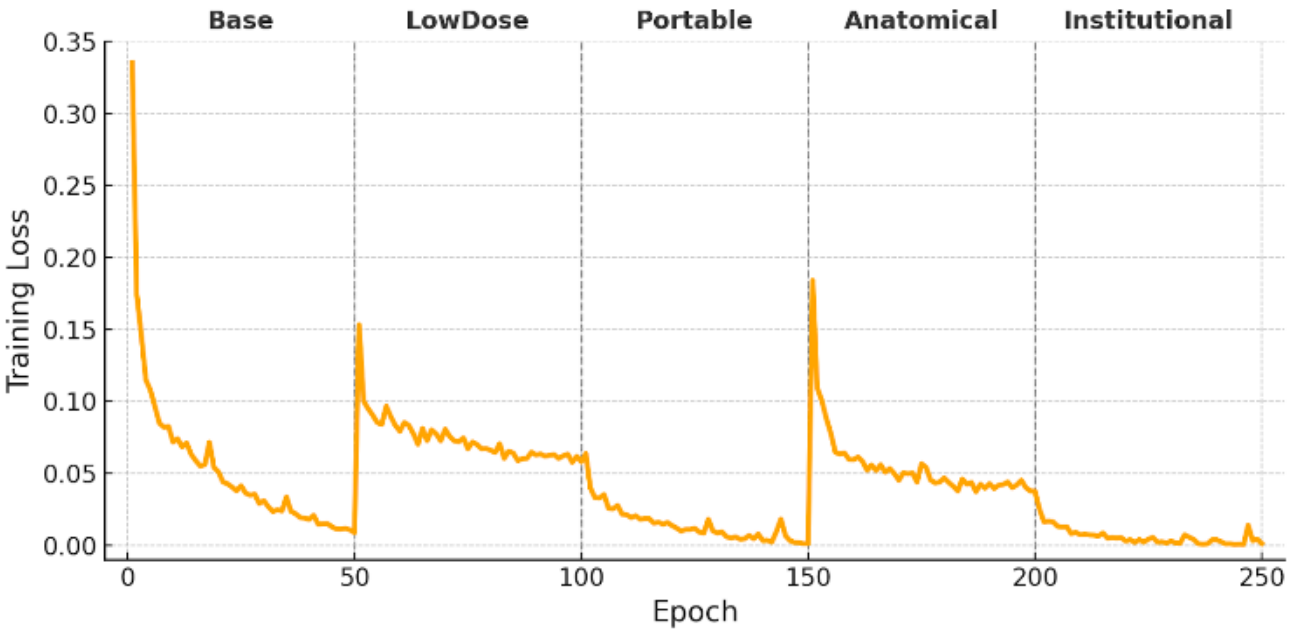

**Fig. 5. Training loss of PneumoNet over five domains (50 epochs each) in Run 1.** (Orange: training loss; gray lines: domain transitions)

Fig. 5 shows the changes in training loss of PneumoNet over 250 epochs for five domain changes (50 epochs per each domain), with domain transitions marked by vertical lines. The training loss is tracked across domains to observe how the model adapts to new data while keeping past knowledge. A stable and decreasing loss indicates effective continual learning without forgetting. In all five domains, the loss starts high but quickly drops below 0.10 within the first few epochs and falls below 0.05 after continued training. It then gradually decreases and stabilizes, reaching nearly zero for the Base, Portable, and Institutional domains. For the LowDose and Anatomical domains, the minimum loss settles around 0.05, showing that these domains are more variable and harder to learn. Notably, the loss does not spike after domain transitions, proving that PneumoNet adapts smoothly to new data while preserving previous knowledge. The consistent decline in loss shows that the combination of dual-stage balanced buffer and dynamic class-weighted loss effectively prevents catastrophic forgetting.

In summary, PneumoNet performs well with small buffers, balanced replay ratio, and different optimizers. It matches MobileNet-V2 in accuracy while being much smaller and faster, showing strong efficiency for continual learning on limited devices.

### 4.4 Efficiency Analysis

Efficiency is especially important for on-device continual learning, where storage and computing power are limited. This section examines whether PneumoNet can keep high accuracy while remaining lightweight and practical for real-time use in clinical environments.

To assess this, we compare PneumoNet with standard CNN and MobileNet-V2 models using key metrics such as model size, parameter count, memory usage, FLOPs, training time, and inference time. These metrics help show how each model balances accuracy and efficiency. In the standard CNN, each stage includes a 3×3 convolution layer, a ReLU activation, and a 2×2 max-pooling layer. The flattened feature map (3,136 units) connects to a 128-unit ReLU layer and two output nodes. All experiments were conducted on Google Colab with an NVIDIA T4 GPU for consistent comparison.

**Table 5. Model architecture comparison of the proposed method based on three other baselines.** (Model Size (MB), Memory Usage (MB), FLOPs (MFLOPs), and Parameters are reported.)

| Model | Model Size (MB) | Memory (MB) | FLOPs (MFLOPs) | Parameters |
|---|---|---|---|---|
| CNN | 1.61 | 1.60 | 135.68 | 420,610 |
| MobileNet-V2 (No-weights) | 8.71 | 8.49 | 197.57 | 2,225,858 |
| MobileNet-V2 (Pretrained) | 8.71 | 8.49 | 197.57 | 2,225,858 |
| **PneumoNet** | **0.22** | **0.21** | **22.60** | **56,194** |

Table 5 shows the efficiency comparison of the different model architectures. PneumoNet is substantially smaller and more efficient than all three baseline architectures. Its model size is reduced to 0.22 MB, roughly seven times smaller than baseline CNN (1.61 MB) and nearly forty-times smaller than MobileNet-V2 (8.71 MB). The memory usage in PneumoNet decreased to 0.21 MB, compared to 1.60 MB in CNN and 8.49 MB in MobileNet-V2. Computationally, PneumoNet requires only 22.6 MFLOPs, a six-fold reduction compared to CNN (135.68 MFLOPs) and nine-fold compared to MobileNet-V2 (197.57 MFLOPs). Furthermore, PneumoNet also uses only 56,194 parameters, fewer than 420K and 2,225K parameters of CNN and MobileNet-V2, respectively. These test results demonstrate the clear suitability of PneumoNet for resource-constrained, real-time continual learning updates on portable devices.

Fig. 6 compares the training times of different models on Base domain. It completes a full training run in 63.52 seconds, marginally faster than the standard CNN (69.30 s) and more than three times faster than both MobileNet-V2 (219.42 s for No-weights and 222.10 s for ImageNet-pretrained). In this test, PneumoNet shows an approximate 8% speedup over CNN and an about 3.5 times acceleration over MobileNet-V2, demonstrating that the proposed model not only minimizes computational and memory overhead but also substantially shortens the model learning cycle. Such rapid convergence is particularly critical for on-device continual learning, where frequent incremental updates must be performed under strict time limits

Fig. 7 compares the inference times of different models for processing a single domain. The inference time evaluation further highlights the model's suitability for real-time, on-device deployment. PneumoNet and CNN achieve inference time of 0.08 s, substantially outperforming MobileNet-V2. Specifically, PneumoNet is approximately 2.3 times faster than the MobileNet-V2 (No-weights, 0.18 s) and nearly 3.8 times faster than the Pretrained MobileNet-V2 (0.30 s). These results demonstrate PneumoNet's ability to provide real-time predictions while remaining lightweight.

In summary, PneumoNet achieves competitive accuracy while being smaller, faster, and more memory-efficient than other models, demonstrating its use for continual learning in resource-limited settings.

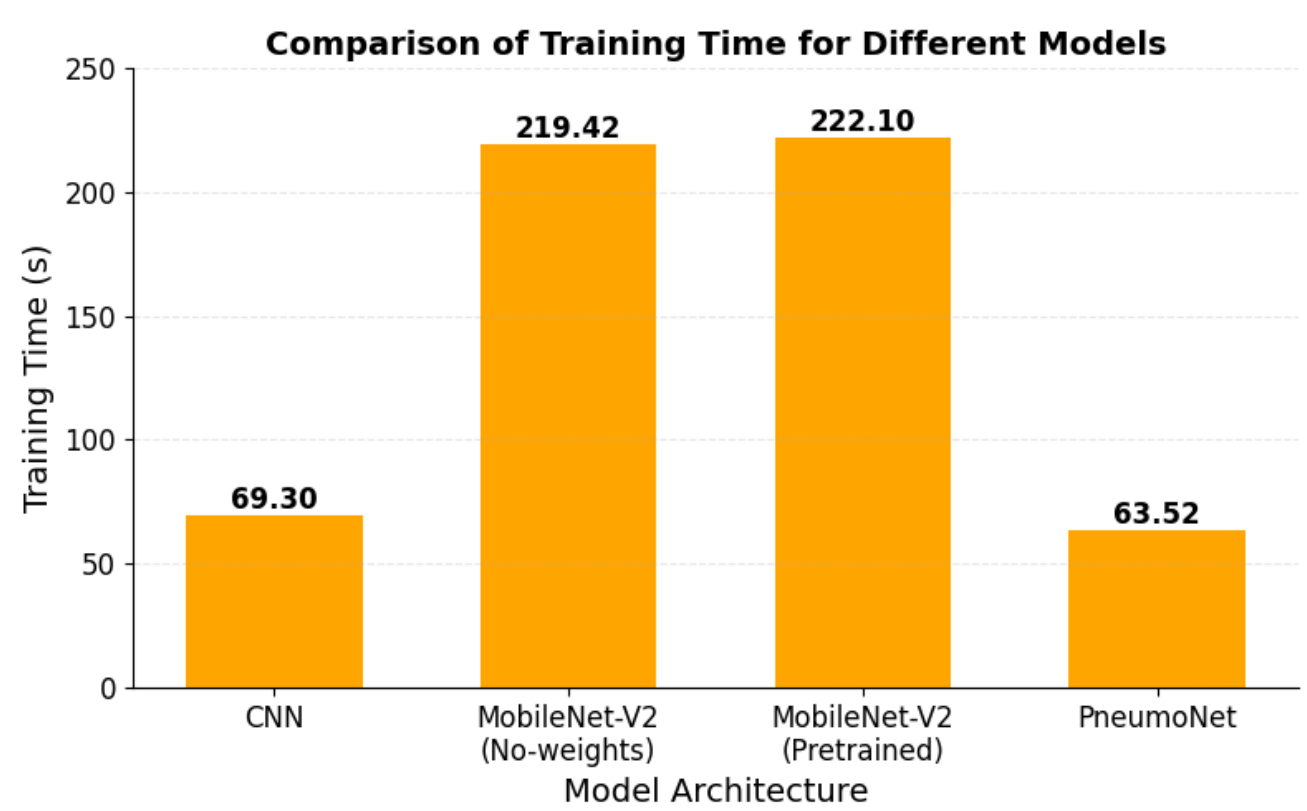

**Fig. 6. Training time comparison of PneumoNet and baselines, on base domain.**

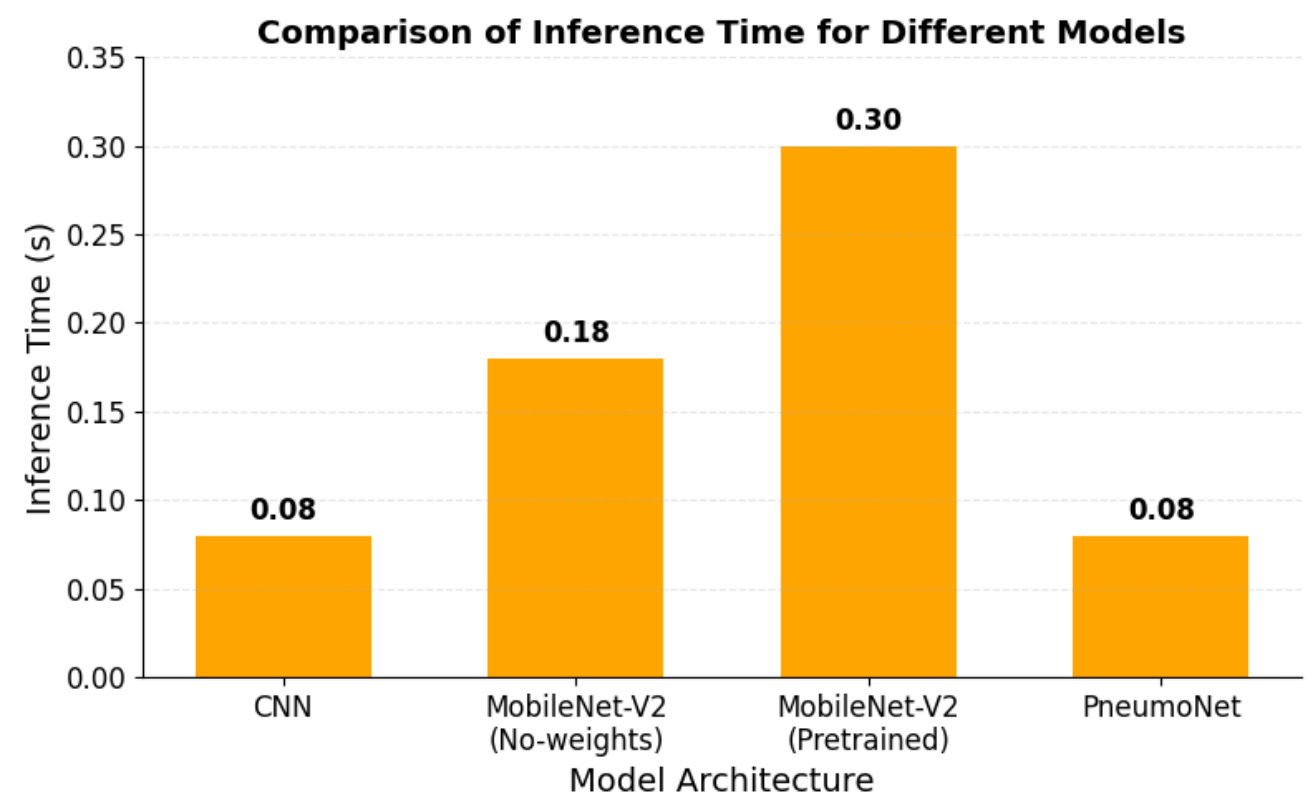

**Fig. 7. Inference time comparison of PneumoNet and baselines, on base domain.**

## 5. CONCLUSION

In this study, we propose PneumoNet, the first domain-incremental continual learning framework for pneumonia detection from chest X-rays. Designed for resource-constrained deployment, PneumoNet integrates a lightweight CNN architecture for on-device prediction, a dual-stage balanced buffer for class-balanced replay, and a dynamic class-weighted loss for adaptive optimization under intra-batch imbalance.

Experiments demonstrate that PneumoNet achieves strong accuracy with minimal forgetting, outperforming state-of-the-art methods while remaining highly efficient. Efficiency analysis confirms its suitability for real-time, resource-limited deployment. Beyond pneumonia detection, these results illustrate the broader potential of lightweight domain-incremental learning for mobile healthcare AI. In future pandemics and other resource-constrained clinical settings, such models enable adaptive, privacy-preserving diagnostic capabilities directly on point-of-care medical devices.

Future work will include evaluating the proposed method on real embedded medical devices, optimizing replay strategies to further improve performance, and validating the model on datasets with naturally occurring domain shifts across multiple institutions.